\documentclass[11pt,a4paper]{article}
\usepackage[hyperref]{conll-2019}
\usepackage{times}
\usepackage{latexsym}
\usepackage{url}
\usepackage{multirow}
\usepackage{graphicx, amsmath,amssymb}
\usepackage[ruled,vlined]{algorithm2e}
\usepackage{booktabs}

\newcommand{\mathbbm}[1]{\text{\usefont{U}{bbm}{m}{n}#1}}
\newcommand{\cev}[1]{\reflectbox{\ensuremath{\vec{\reflectbox{\ensuremath{#1}}}}}}

\aclfinalcopy 

\title{Unsupervised Adversarial Domain Adaptation for Implicit Discourse Relation Classification}

\author{Hsin-Ping Huang \\
  Department of Computer Science \\
  The University of Texas at Austin \\
  {\tt hsinping@cs.utexas.edu} \\\And
  Junyi Jessy Li \\
  Department of Linguistics \\
  The University of Texas at Austin \\
  {\tt jessy@austin.utexas.edu} \\}

\date{}

\begin{document}
\maketitle
\begin{abstract}
   Implicit discourse relations are not only more challenging to classify, but also to annotate, than their explicit counterparts. We tackle situations where training data for implicit relations are lacking, and exploit domain adaptation from explicit relations~\cite{ji2015closing}. We present an unsupervised adversarial domain adaptive network equipped with a reconstruction component. Our system outperforms prior works and other adversarial benchmarks for unsupervised domain adaptation. Additionally, we extend our system to take advantage of labeled data if some are available. 
\end{abstract}

\section{Introduction}

Discourse relations capture the relationship between units of text---e.g.,  sentences and clauses---and are an important aspect of text coherence. While some relations are expressed explicitly with a discourse connective (e.g., ``for example'', ``however''), relations are equally often expressed implicitly without an explicit connective~\cite{Prasad08thepenn}; in these cases, the relation needs to be inferred.

Resources for implicit discourse relations are scarce compared to the explicit ones, since they are harder to annotate~\cite{miltsakaki2004penn}. For example, among corpora annotated with discourse relations such as Arabic~\cite{al2010leeds}, Czech~\cite{polakova2013introducing}, Chinese~\cite{zhou2015chinese}, English~\cite{Prasad08thepenn}, Hindi~\cite{oza2009hindi}, and Turkish~\cite{zeyrek2013turkish}, only the Chinese, English and Hindi corpora include implicit discourse relations~\cite{prasad2014reflections}. In this low-resource scenario, \citet{ji2015closing} proposed training with explicit relations via unsupervised domain adaptation, viewing explicit relations as a source domain with labeled training data, and implicit relations as a target domain with no labeled data. The domain gap between explicit and implicit relations is acknowledged by prior observations that the two types of discourse relations are linguistically dissimilar~\cite{sporleder2008using,rutherford15}.

We present a new system for the unsupervised domain adaptation setup on the Penn Discourse Treebank~\cite{Prasad08thepenn}. Our system is based on Adversarial Discriminative Domain Adaptation~\cite{tzeng2017adversarial}, 
which decouples source domain training and representation mapping between source and target. 
We improve this framework by proposing a reconstruction component to preserve the discriminability of target features, and incorporating techniques for stabler training on textual data.

Experimental results show that even with a simple architecture for representation learning, our unsupervised domain adaptation system outperforms prior work by 1.4-2.3 macro F1, with substantial improvements on Temporal and Contingency relations. It is also superior to DANN~\cite{Ganinjmlr}, an adversarial framework widely used in NLP~\cite{Chen2016AdversarialDA,Gui2017PartofSpeechTF,aspect,Fu2017DomainAF,Joty2017CrosslanguageLW,cross17}, by 5.7 macro F1.

Finally, we extend the system to incorporate in-domain supervision as it is sometimes feasible resource-wise to build a seed corpus that may not be large enough to train a fully supervised system. We simulate this scenario by enabling the system to jointly optimize over a varying number of labeled examples of implicit relations. Our system consistently outperforms two strong baselines.

\section{Related Work}\label{sec:related}

\citet{sporleder2008using} and \citet{rutherford15} observed that explicit and implicit relations are linguistically dissimilar, warranting an unsupervised domain adaptation approach in ~\citet{ji2015closing}. They  
used a marginalized denoising autoencoder to obtain generalized feature representations across the source and target domains with a linear SVM as the classification model. Our system improves upon this work using an adversarial network; we further generalize our network to semi-supervised settings.

To supplement the training data of implicit discourse relations, prior works have used weak supervision from sentences with  discourse connectives~\cite{marcu2002unsupervised,sporleder2008using,braud14,ji2015closing}, by analyzing connectives ~\cite{zhou10_2,zhou10,biran2013aggregated,rutherford15,braud16,wu17}, using a multi-task framework with other corpora~\cite{lan13,Liu:2016,Lan2017MultitaskAN}, or utilizing cross-lingual data~\cite{wu16,shi17}. The important distinction between this work and the research above is that these are supervised systems that used all of the annotated {\em implicit} annotation from PDTB during training, while exploring non-PDTB corpora for additional, noisy discourse cues; on the contrary, our main goal is to assume no labeled training data for implicit discourse relations.

Unsupervised domain adaptation with adversarial networks has become popular in recent years; this type of approach  
generates a representation for the target domain with the goal that the discriminator unable to distinguish between the source and target domains. Prior works proposed both generative approaches~\cite{liu16,bousmalis17,Swami17,Russo17} and discriminative approaches~\cite{Ganinjmlr,Tzeng15,tzeng2017adversarial}. The discriminative DANN algorithm from \citet{Ganinjmlr} is frequently used in NLP tasks~\cite{Chen2016AdversarialDA,Gui2017PartofSpeechTF,aspect,Fu2017DomainAF,Joty2017CrosslanguageLW,cross17}. 
Our method builds upon Adversarial Discriminative Domain Adaptation 
~\cite{tzeng2017adversarial}, shown to outperform DANN in visual domain adaptation but has not been used in NLP tasks. The key differences between the two are discussed in Section~\ref{sec:model}.

\citet{qin17} adopted adversarial strategies to supervised implicit discourse classification. They train an adversarial model using implicit discourse relations with and without expert-inserted connectives. Note again that theirs is a fully supervised system using signals in addition to the implicit relation annotations themselves, while our main focus is unsupervised domain adaptation that does not train on implicit relations.

\section{Model Architecture}\label{sec:model}

\begin{figure*}[t]
\centering
\includegraphics[width=0.9\linewidth]{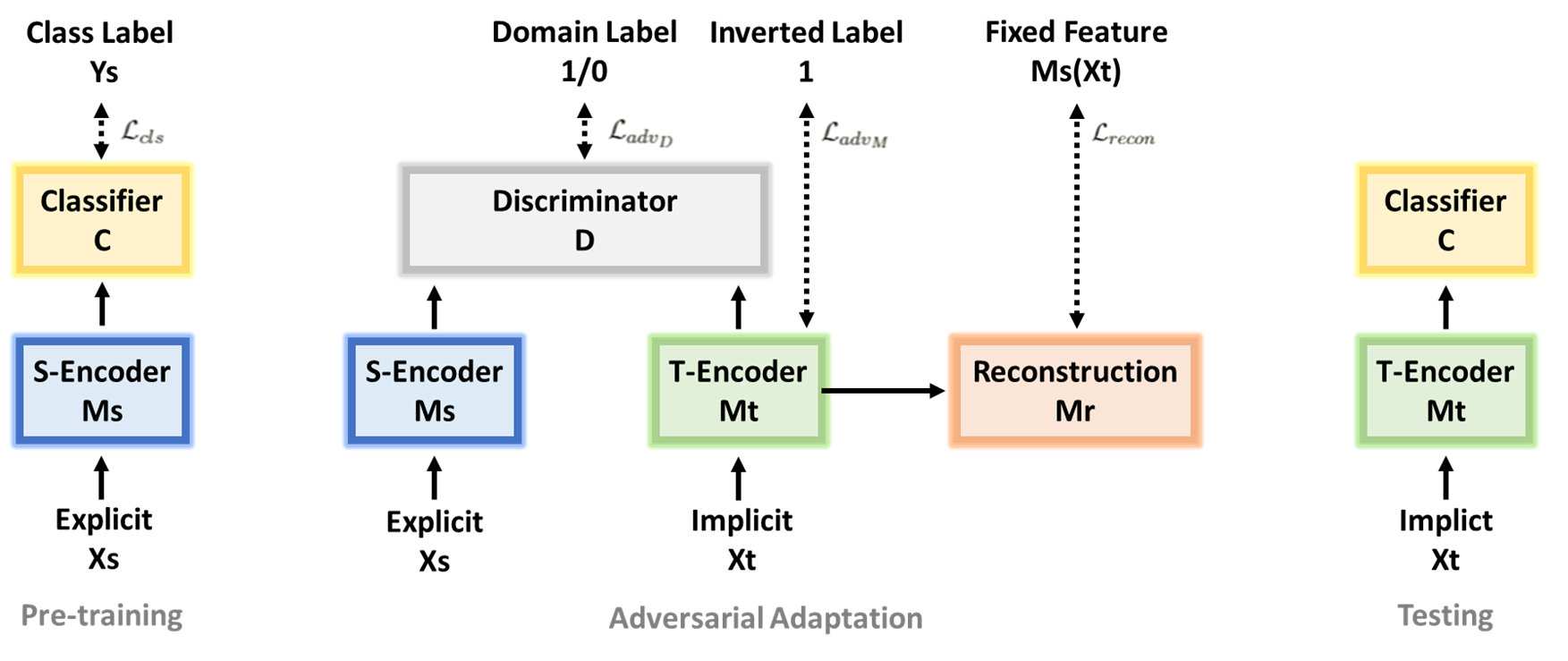}
   \caption{The framework of our proposed adversarial domain adaptation model, containing the pre-training stage, the adversarial adaptation stage, and the testing stage. The dashed box shows the supervised component.}
\label{fig:structure}
\end{figure*}

To classify discourse relations, our system takes a pair of sentence arguments $x$ as input, and outputs the discourse relation $y$ between these two arguments. 
With unsupervised domain adaptation, we have examples $(X_s, Y_s)$  from the source domain, i.e., explicit discourse relations, and unlabeled examples $(X_t)$ from a target domain, i.e., implicit discourse relations. 

We use ADDA~\cite{tzeng2017adversarial} as our underlying framework for domain adaptation. 
ADDA first learns a discriminative representation for the  classification task in the source domain, then learns a representation for the target domain that mimics the distribution of the source domain. The key insight here is  asymmetric mapping, where the target representation is ``updated'' until it matches with the source, a process more similar to the original Generative Adversaial Networks~\cite{goodfellow2014generative} than joint training as in DANN~\cite{Ganinjmlr}. 
Intuitively, since ADDA learns distinct feature encoders for the source and target domains instead of using a shared encoder, the same network doesn't have to handle instances from different domains. 

Summarized in Figure~\ref{fig:structure}, we first pre-train a source encoder $M_s$ and source classifier $C$ (Section~\ref{sec:base}), then train the target encoder $M_t$ (initialized with $M_s$) and discriminator $D$ in an adversarial way, to minimize the domain discrepancy distance between the target representation distribution $M_t$($X_t$) and that of the source $M_s$($X_s$) (Section~\ref{sec:adda}). Eventually, the target feature space is trained to match the source,  
and the source classifier $C$ can be directly used on the target domain.

\subsection{Base encoder and classifier}\label{sec:base}

The source and target {\bf encoders} $M_s$ and $M_t$ follow the same architecture; $M_t$ is initialized to be $M_s$ during adaptation. The encoders encode relation arguments into latent representations, and then feeds the representations into a {\bf classifier $C$} to predict the discourse relation.

\begin{figure}[t] 
\centering
\includegraphics[width=0.8\linewidth]{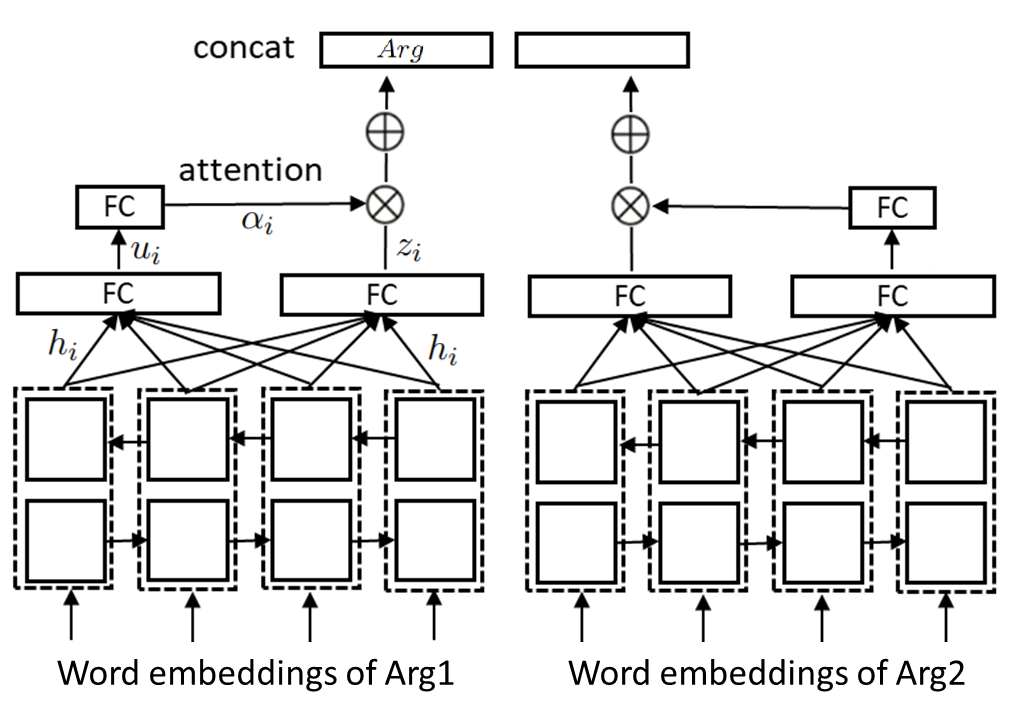}
   \caption{Neural structure of inner-attention BiLSTM to encode relation argument pairs.}
\label{fig:encoder}
\end{figure}

\paragraph{Encoder} The encoder generates a representation for each argument with an inner-attention Bidirectional LSTM \cite{Yang2016HierarchicalAN} shared between the two arguments. Then, the representations of the two arguments are concatenated to form the final representation, shown in Figure \ref{fig:encoder}.

Specifically, we encode each word in an argument into its word embeddings, which are fed into a BiLSTM, to get the hidden representations $z_i$ using a fully-connected layer $W_c$ on top of the concatenated hidden states $h_i=[\vec{h}_{i}, \cev{h}_{i}]$. 
We then apply an attention mechanism to induce a distribution of weights over all tokens in the argument; the final argument representation $Arg$ is a 
weighted sum of $z_i$ based on the attention weights $\alpha_i$: 
\begin{align}
z_i &=W_ch_i+b_c \nonumber \\
u_i &=tanh({W_w}{h_i}+b_w) \nonumber \\
\alpha_i &=\frac{exp(u_i^Tu_w)}{\sum_{i}exp(u_i^Tu_w)} \nonumber \\
Arg&=\sum_{i}\alpha_iz_i
\end{align}
Where $W_c, b_c, W_w, b_w, u_w$ are model parameters.

\paragraph{Classifier} The classifier consists of a single fully-connected layer on top of the encoder, finished with a softmax classification layer.

The source encoder $M_s$ and the classifier $C$ are trained using a supervised loss:
\begin{align} \label{eq:1}
\min_{M_s, C}\mathcal{L}_{cls}(X_s, Y_s) &= \nonumber \\ 
\mathbb{E}_{(x_s,y_s)}-\sum_{k}\mathbbm{1}&[k=y_s]\log{C(M_s(x_s))}
\end{align}

\subsection{Unsupervised adversarial domain adaptation}\label{sec:adda}
We then learn a target encoder $M_t$ to generate features for the target data which can be classified with classifier $C$, without assuming labels $Y_t$ in the target domain. This is achieved by training 
a domain discriminator $D$, which classifies whether a feature is from the source or the target domain, and the target encoder $M_t$, that produces features similar to the source domain features and tries to fool the discriminator to predict the incorrect domain label.

The discriminator $D$ is optimized according to a standard supervised loss:
\begin{align} \label{eq:2}
\min_{D}\mathcal{L}_{adv_D}&(X_s, X_t, M_s, M_t) = \nonumber \\
&-\mathbb{E}_{x_s}[\log{D(M_s(x_s))}] \nonumber \\
&-\mathbb{E}_{x_t}[\log{(1-D(M_t(x_t)))}]
\end{align}
$D$ consists of two fully-connected layers on top of the encoder, finished with a softmax classification layer.

The target encoder $M_t$ is optimized according to a standard GAN loss with inverted labels:
\begin{align} \label{eq:3}
\min_{M_t}\mathcal{L}_{adv_M}(X_s, X_t, D) &= \nonumber \\
-\mathbb{E}_{x_t}&[\log{D(M_t(x_t))}]
\end{align}

\paragraph{Spectral normalization} To stabilize the training of the discriminator, we employ spectral normalization, a weight normalization technique 
~\cite{miyato2018spectral}, which controls the Lipschitz constant of the discriminator function by constraining the spectral norm of each layer. Spectral normalization is easy to implement without tuning any hyper-parameters and has a small additional computational cost.

\paragraph{Label smoothing}
We utilize label smoothing~\cite{szegedy2016rethinking} to regularize the classifier during pre-training, which prevents the largest logit from becoming much larger than all others, and therefore prevents overfitting and makes the classifier, trained in the source domain, more adaptable.

For a source domain training example $x_s$ with ground-truth label $y_s$ and ground-truth distribution $q(k|x_s)$, the classifier computes the classification probability over relation classes as $p(k|x_s)$ for $k\in\left \{1 . . . K\right \}$. With label smoothing, we replace the ground-truth label distribution $q(k|x_s)$ in the standard cross-entropy loss as a linear combination of $q(k|x_s)$ and a uniform distribution over classes $u(k)=1/K$.

\begin{align} \label{eq:5}
\min_{M_s, C}\mathcal{L}_{cls}(X_s, Y_s) &= -\sum_{k}q'(k)\log{(p(k))} \nonumber \\
q'(k|x_s)=(1-\epsilon)&\mathbbm{1}[k=y_s]+\epsilon/K
\end{align}

\paragraph{Reconstruction loss}

In order to classify the target representations using the source classifier, the target encoder is trained to produce representations that mimic the source domain representations in the adversarial training stage. Since there is no supervised loss applied in this stage, the target encoder may lose its ability to produce discriminative features that are helpful during classification. We propose a reconstruction loss to preserve the discriminability of the target encoder when adversarially adapting its features.

\begin{figure}[t]
\centering
\includegraphics[width=0.7\linewidth]{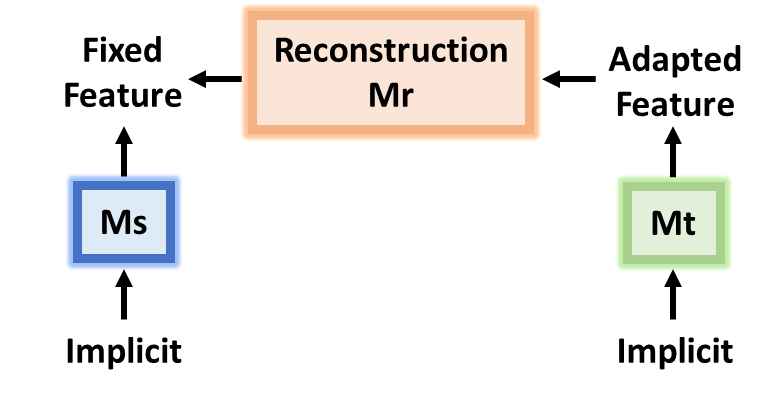}
   \caption{The reconstruction loss component augmenting our unsupervised adversarial domain adaptor.}
\label{fig:mapping}
\end{figure}

Since we initialize the target encoder with the source encoder, the initial representation (before domain adaptation) of a target instance $x_t$ is the representation of {\em target} instances produced by the {\em source} encoder $M_s(x_t)$ (which is then fixed). After training, $M_t(x_t)$ adapts to the source domain and becomes dissimilar to $M_s(x_t)$.  The reconstruction loss encourages the target encoder to produce features that can be reconstructed back to $M_s(x_t)$
(Figure \ref{fig:mapping}).

For a target example $x_t$, we learn a reconstruction mapping $M_r$ that maps the target representation $M_t(x_t)$ to $M_s(x_t)$:
\begin{equation} \label{eq:6}
x_t \rightarrow M_t(x_t) \rightarrow M_r(M_t(x_t)) \approx M_s(x_t)
\end{equation}
The target encoder $M_t$ and the reconstruction mapping $M_r$ are optimized jointly with a reconstruction loss:
\begin{align} \label{eq:7}
\min_{M_t, M_r}\mathcal{L}_{recon}(X_t, M_s) =& \nonumber \\ 
\mathbb{E}_{x_t}[\lvert\lvert M_r(M_t&(x_t))-M_s(x_t)\rvert\rvert_{2}^{2}]
\end{align}
$M_r$ consists of three fully-connected layers on top of the encoder.

\paragraph{Unsupervised objective}

For unsupervised domain adaptation, our full objective is:
\begin{align} \label{eq:8}
\mathcal{L}^{unsup}(X_s, Y_s, &X_t, M_s, M_t, D) = \nonumber \\
&\min_{M_s, C}\mathcal{L}_{cls}(X_s, Y_s) + \nonumber \\
&\min_{D}\mathcal{L}_{adv_D}(X_s, X_t, M_s, M_t) + \nonumber \\
&\min_{M_t}\mathcal{L}_{adv_M}(X_s, X_t, D) + \nonumber \\
&\min_{M_t, M_r}\mathcal{L}_{recon}(X_t, M_s)
\end{align}

\subsection{Training}

\begin{algorithm}[t] \label{alg}
\small
\KwIn{ explicit sentences with labels \{$x_s, y_s$\} \\
~~~~~~~~~~~~~implicit sentences without labels \{$x_t$\} }
\SetKwInOut{Parameter}{Notations}
\Parameter{source encoder $M_s$, classifier $C$,\\
target encoder $M_t$,\\
reconstruction mapping $M_r$,\\
domain discriminator $D$}

\nl Train $M_s, C$ through Eq.(\ref{eq:1}) with $\{x_s, y_s\}$\;

\nl Initialize $M_t$ as $M_s$\;
\nl {\bf Repeat}\\

\nl ~~~Train $D$ through Eq.(\ref{eq:2}) with $\{x_s, x_t\}$\ and train $M_t$ through Eq.(\ref{eq:3}) with $\{x_t\}$\;

\nl ~~~Train $M_t, M_r$ through Eq.(\ref{eq:7}) with $\{x_t\}$\;

\KwOut{$M_t$ and $C$ for relation prediction} 

    \caption{{Adversarial Adaptation} \label{Algorithm}}
\end{algorithm}

Summarized in Algorithm \ref{Algorithm}, the training procedure consists of three stages: pre-training, adversarial adaptation, and testing. During pre-training, we train the source encoder $M_s$ and the classifier $C$ according to Eq.(\ref{eq:1}). 
In the adversarial adaptation stage, we alternately train the discriminator $D$, target encoder $M_t$, and reconstruction mapping $M_r$ according to Eq.(\ref{eq:2}), Eq.(\ref{eq:3}), Eq.(\ref{eq:7}).
Finally, we test the model using the target encoder $M_t$ and classifier $C$. The steps (lines 4 and 5) in the Repeat loop execute once in one iteration, and we optimized the model in two-step units.

\section{Unsupervised Domain Adaptation Experiments}

We first evaluate our model for the default task: unsupervised domain adaptation from explicit discourse relations to implicit discourse relations.

\subsection{Settings}

\paragraph{Data} We train and test our model on the PDTB,  following the experimental setup of \citet{ji2015closing}. The test examples are implicit relation instances from PDTB sections 21-22. The explicit training set consists of explicit examples from sections 02-20 and 23-24, and the explicit development set consists of the explicit examples from sections 00-01. The implicit training set and the implicit development set consist of examples from the same sections as the explicit sets. Evaluation is done for the first-level relations---Temporal, Comparison, Contingency, and Expansion.  Table 
\ref{tb:statistic} summarizes the statistics of the four top-level implicit and explicit discourse relations in the PDTB.

\paragraph{Training Details}

We early-stopped training for both stages before total convergence if the macro F1 on the development set does not improve. 
During pre-training (Line 1 in Algorithm~\ref{Algorithm}), we train and validate the model on the explicit training and development set. Early stopping happened after around 20 epochs. 
During adversarial adaptation (lines 4 and 5 in Algorithm 1), we train the model on the explicit and implicit training sets {\em without relation labels $Y_t$}, and validate on the implicit development set\footnote{Using a development set in the target domain is common in unsupervised domain adaptation \cite{Ganinjmlr,liu16,tzeng2017adversarial,bousmalis17,Russo17} }. Early stopping happened after around 5 epochs (with lines 4 and 5 executed once in each epoch).

\paragraph{Model configuration}

The hyperparameters, as well as the number of fully connected layers for the classifier $C$, discriminator $D$ and the reconstruction mapping $M_r$, are all set according to the performance on the development sets. We first set the hyper-parameters of the encoders $M_s, M_t$ and classifier $C$ based on development performance during the pre-training stage. Then, we set the hyper-parameters of $D$ and $M_r$ based on development performance of the adaptation stage.

We use GloVe \cite{pennington2014glove} for word embeddings with dimension 300.  The maximum argument length is set to 80. The encoder contains an inner-attention BiLSTM with dimension 50, producing a representation with dimension 200 for each example. The discriminator $D$ consists of 2 hidden layers with 200 and 200 neurons on each layer. The reconstruction mapping $M_r$ contains 3 hidden layers with 120, 15 and 120 neurons on each layer. The label smoothing parameter $\epsilon$ is 0.1. 
We use Adam~\cite{Kingma2014AdamAM}  with learning rate 1e-4 for the base encoder and classifier, and 1e-6 for the adversarial domain adapter. We use SGD optimizer with learning rate 1e-2 for the reconstruction component.
All the models were implemented using PyTorch~\cite{paszke2017automatic} and adapted from \citet{conneau2017supervised}.

\begin{table} 
\centering
\small
\begin{tabular}{ l|c|c|c|c|c } 
& \multicolumn{2}{|c|}{Explicit} & \multicolumn{3}{|c}{Implicit} \\ \midrule
Relation & Train & Dev & Train & Dev & Test \\ \toprule
Temporal    & 2904 & 288 & 704 & 68 & 54\\
Contingency & 2792 & 181 & 3622 & 276 & 287\\
Comparison  & 4674 & 366 & 2104 & 146 & 191\\
Expansion   & 5342 & 450 & 7394 & 556 & 651 \\ \bottomrule
\end{tabular}
   \caption{The number of examples of the four top level discourse relations in PDTB 2.0.}
   \label{tb:statistic}
\end{table}

\begin{table*}[t]
\centering
\small
\begin{tabular}{ l|cccc|c } 
&Temporal&Contingency&Comparison&Expansion&Macro $F_1$\\ \toprule
Implicit$\,\to\,$Implicit & 25.53 & 41.02 & 30.35 & 65.38 & 40.57 \\ \midrule
Explicit$\,\to\,$Implicit & 22.22 & 22.35 & 23.06 & 57.86 & 31.37 \\
+Domain Adaptation & 30.62 & 42.71 & 25.00 & 52.23 & 37.64 \\
~~~+Spectral Normalization & 30.20 & 45.42 & 21.90 & 58.72 & 39.06 \\
~~~~~~+Label Smoothing & {\bf 31.58} & 46.40 & 24.64 & 58.03 & 40.16 \\

~~~~~~~~~+Reconstruction & 31.25 & {\bf 48.04} & 25.15 & 59.15 & {\bf 40.90} \\ \midrule
\cite{ji2015closing} & 19.26 & 41.39 & 25.74 & 68.08 & 38.62\\
~+weak supervision & 20.35 & 42.25 & {\bf 26.32} & {\bf 68.92} & 39.46\\ \midrule

DANN & 26.19 & 34.20 & 25.74 & 54.70 & 35.21\\

\bottomrule
\end{tabular}
   \caption{Per-class and macro average F1 (\%) of unsupervised domain adaptation from explicit to implicit relations.}
\label{tab:da}
\end{table*}

\subsection{Systems}

We experiment with three settings:
   
\paragraph{Implicit $\rightarrow$ Implicit} A supervised implicit discourse relation classifier using the base encoder and classifier, optimizing the standard cross-entropy loss, using the full implicit training and development sets. This model does not use the explicit relations.

\paragraph{Explicit $\rightarrow$ Implicit} A discourse relation classifier using the explicit training set and implicit development set, optimizing the standard cross-entropy loss. This serves as a baseline without any domain adaptation.

\paragraph{Domain adaptation} Our full adaptation model trained on the explicit training set, and adapted to the implicit training set without relation labels. We perform ablation study with different extensions describe in Section \ref{sec:model}: spectral normalization, label smoothing, and reconstruction loss.

\vspace{0.5em}
For {\em benchmarking}, we train an unsupervised domain adaptation system  using {\bf DANN}~\cite{Ganinjmlr}, which jointly learns domain-invariant representations and the classifier and is often used in NLP (c.f.\ Section~\ref{sec:related}). We use the same encoder, classifier and discriminator structures, with parameters tuned on the implicit development data. The system is optimized using Adam with learning rate 2e-4 and the adaptation parameter 0.25, chosen between 0.01 and 1 on a logarithmic scale.

\subsection{Results}

To evaluate our model, we train four-way classifiers and report \textbf {per-class and macro F1 scores}. 
Table~\ref{tab:da} tabulates the experimental results for unsupervised domain adaptation. 

We also show reported results from \citet{ji2015closing}. Even though they trained four binary classifiers (instead of doing multi-class classification), it is the only prior work exploring unsupervised domain adaptation for implicit discourse relation classification. 
We include two settings: their best system with labeled data from PDTB explicit relations only (and an implicit development set), and their system with additional weak supervision from non-PDTB sources.\footnote{The weakly labeled data includes sentences extracted from 1000 CNN articles, with explicit discourse connectives but without annotated discourse relations.}

Our full system achieves the best average F1 measure, a 9.53\% absolute increase from Explicit$\,\to\,$Implicit. It also performs 2.28\% better than \citet{ji2015closing}'s model trained without weak supervision, and 1.44\% better than their model trained with weak supervision. The full system achieved an average F1 comparable to the supervised Implicit$\,\to\,$Implicit, while \citet{ji2015closing}'s models did not. Comparing with DANN, our system achieved superior performance for 3 of the 4 relations, showing that training target representations and the classifier in two stages outperforms doing both jointly.

The largest improvements from the Explicit$\,\to\,$Implicit baseline are from Temporal (from 22.22 to 31.25) and Contingency (from 22.35 to 48.04) relations. Our system performs $\sim$11\% better for Temporal, and $\sim$6\% better for Contingency, than \citet{ji2015closing}'s binary classifiers. The Comparison and Expansion relations improved by about 2\% from the baseline, a smaller improvement compared to the other two relations. Our Comparison performance is comparable with \citet{ji2015closing}'s model without weak supervision.

Notably, the performance for Expansion dropped after domain adaptation (without extensions) by about 5\%. We suspect that this is because  
the distributions of Expansion 
among other relations are very different (33\% for explicit and 53\% for implicit, c.f. Table~\ref{tb:statistic}). By applying Spectral Normalization, the performance improved and surpassed Explicit$\,\to\,$Implicit.

Component-wise, Spectral Normalization helps two of the four relations (Contingency and Expansion), but hurts the performance of Comparison. Label smoothing improves performance for all relations except Expansion; applying the reconstruction loss improves performance for all relations except Temporal. Overall, the best result on this task is to incorporate all components.

\begin{figure}[t]
\centering
\includegraphics[width=0.8\linewidth]{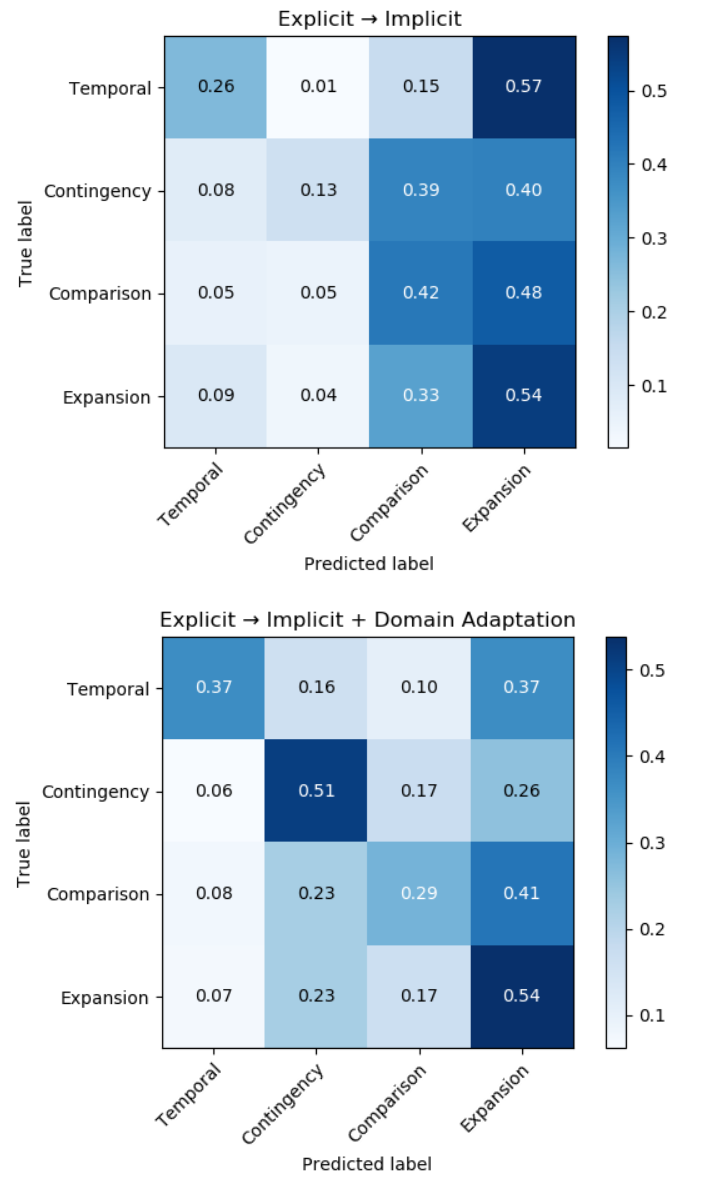}
   \caption{Normalized confusion matrices before and after unsupervised domain adaptation.}
\label{fig:confusion}
\end{figure}

\paragraph{Error analysis} Figure \ref{fig:confusion} shows the normalized confusion matrices before and after unsupervised domain adaptation. 
Before adaptation, Temporal and Contingency relations are often misclassified as Expansion, which is substantially improved after adaptation. The improvement in F score for Comparison is milder due to lower precision and higher recall, which is also reflected in the matrices. Finally, the drop of performance in Expansion adaptation can be traced through increased confusion between Expansion and Contingency.

\section{How about a little supervision?}

We have so far presented an unsupervised domain adaptation system that is not trained on any labels $Y_t$ in the target domain. However, it is sometimes feasible to have some seed annotation that can be used to improve prediction. 
Hence we extend the model with an optional supervised component. We evaluate this extension by gradually adding  labeled examples of implicit discourse relations, simulating situations when different numbers of labeled examples are available.

\subsection{Incorporating supervision}\label{sec:supmodel}

\begin{figure}[t]
\centering
\includegraphics[width=0.65\linewidth]{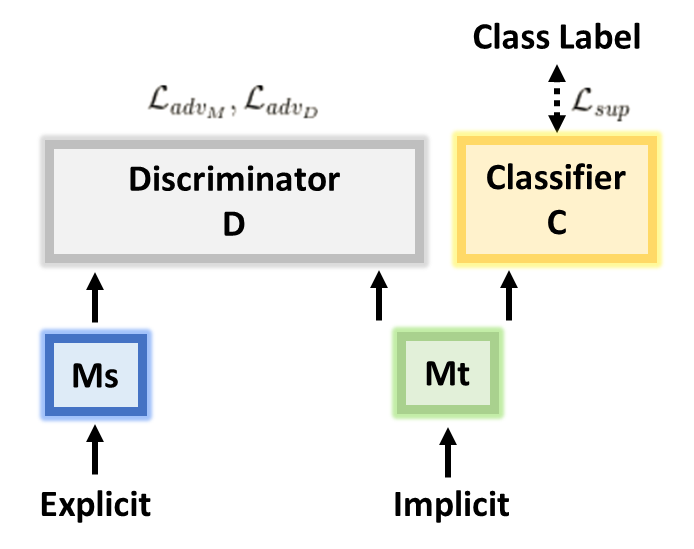}
   \caption{An extension component to incorporate supervision with our unsupervised adversarial domain adaptor.}
\label{fig:supervised}
\end{figure}

We extend the model with a supervised component, where a subset $X^L_t\subseteq X_t$ has labels $Y^L_t$. 
Illustrated in Figure \ref{fig:supervised}, we jointly optimize the target encoder $M_t$ and the classifier $C$ according to an additional supervised loss:
\begin{align} \label{eq:9}
&\min_{M_t, C}\mathcal{L}_{sup}(X^L_t, Y^L_t) = \nonumber \\
&\mathbb{E}_{(x^L_t,y^L_t)}-\sum_{k}\mathbbm{1}[k=y^L_t]\log{C(M_t(x^L_t))}
\end{align}

Effectively, we encourage the target encoder to jointly extract more discriminative features for all target examples $(X_t)$, and learning target domain representations close to the source.

The full objective incorporates supervision of the in-domain labels by adding $\mathcal{L}_{sup}(X^L_t, Y^L_t)$ to the unsupervised objective:
\begin{align} \label{eq:sup}
\mathcal{L}^{sup}(&X_s, Y_s, X_t, M_s, M_t, D) = \nonumber \\
&\mathcal{L}^{unsup}(X_s, Y_s, X_t, M_s, M_t, D) + \nonumber \\
&\min_{M_t, C}\mathcal{L}_{sup}(X^L_t, Y^L_t)
\end{align}

\subsection{Data and settings}

We synthesize the labeled target subset $(X^L_t, Y^L_t)$ $(X^L_t\subseteq X_t)$ by randomly extracting subsets from the implicit training set and get their labels. The sizes of this subset range from 1382 to 13824 with a stepsize of 1382. Note that we use the entire implicit training set ($X_t$ without relation labels $Y_t$) in the adversarial adaptation process as unlabeled data in the target domain, and the sampled labeled data is used in the supervised component only. 
We use the same hyper-parameters as the unsupervised experiment, except that we tune the learning rate on the implicit development set. 

\subsection{Systems}

We compare three settings:

\paragraph{Supervised baseline}
   The encoder and classifier trained on the sampled implicit instances $(X^L_t, Y^L_t)$, optimizing Eq.(\ref{eq:9}).
   
\paragraph{Pre-training baseline}
   Our model with the supervised component, but without the domain adaptation component. This setting is equivalent to pre-training on the explicit instances then fine-tuning on the sampled implicit instances. This is trained on the explicit training set $(X_s, Y_s)$, plus the sampled implicit instances $(X^L_t, Y^L_t)$, optimizing Eq.(\ref{eq:5}) and Eq.(\ref{eq:9}).

\paragraph{Semi-supervised domain adaptation}
   Our full model with both the supervised and adaptation components, optimizing Eq.(\ref{eq:sup}). The supervised component uses the sampled implicit instances $(X^L_t, Y^L_t)$ for training.

\begin{figure}[t]
\centering
\includegraphics[width=1.0\linewidth]{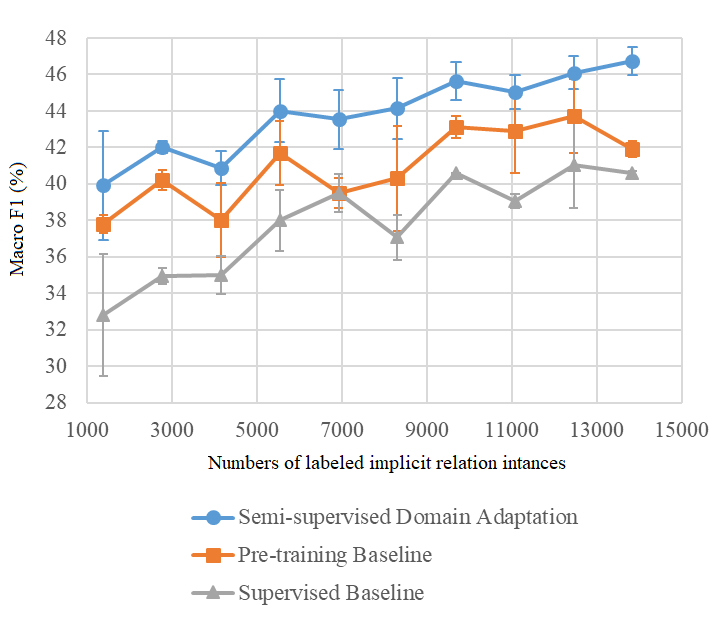}
\vspace{-0.8cm}
\caption{The average F1 (\%) with varying numbers of labeled implicit relation training data.}
\label{fig:onecol}
\end{figure}

\subsection{Results}

Since the added training data is randomly sampled, we average the performance across 3 different runs. Figure~\ref{fig:onecol} shows the average F1 measure (y-axis) of the above three supervised systems, with varying numbers of labeled implicit relation training data (x-axis). Standard errors are also shown in the graph.

Our full system outperforms both the supervised baseline and the pre-training baseline, regardless of the amount of labeled target data. This evaluation also reveals that the pre-training baseline also improves upon the supervised baseline across the board, which means that the performance of implicit relation classification can be improved with pre-training on explicit relations. 

Finally, the macro F1 of our system using full supervision  is 47.50. Since we focus on domain adaptation and used very simple encoders, we do not attempt to achieve state-of-the-art (e.g.,~\citet{dai-huang:2018:N18-1},~\citet{bai-zhao:2018:C18-1}). However this performance is on-par with many recent work using multi-task or GANs, including \citet{Lan2017MultitaskAN} (47.80), \citet{qin17} (44.38, Reproduced results on four-way classification), and \citet{Liu:2016} (44.98). These results confirm that our framework generalizes well with respect to the amount of supervision in the target domain.

\section{Conclusion}
Our work tackles implicit discourse relation classification in a low resource setting that is flexible to the amount of supervision. 
We present a new system based on the adversarial discriminative domain adaptation framework \cite{tzeng2017adversarial} for unsupervised domain adaptation from explicit discourse relation to implicit discourse relation. We propose a reconstruction loss to preserve the discriminability of features during adaptation, and we generalize the framework to make use of possibly available seed data by jointly optimizing it with a supervised loss. Our system outperforms prior work and strong adversarial baselines on unsupervised domain adaptation, and works effectively with varying amount of supervision.

\section*{Acknowledgments}
We thank Ray Mooney, Eric Holgate, and the anonymous reviewers for their helpful feedback. This work was partially supported by the NSF Grant IIS-1850153.

\bibliography{conll-2019}

\begin{thebibliography}{47}
\expandafter\ifx\csname natexlab\endcsname\relax\def\natexlab#1{#1}\fi

\bibitem[{Al-Saif and Markert(2010)}]{al2010leeds}
Amal Al-Saif and Katja Markert. 2010.
\newblock The leeds arabic discourse treebank: Annotating discourse connectives
  for arabic.
\newblock In \emph{LREC}.

\bibitem[{Bai and Zhao(2018)}]{bai-zhao:2018:C18-1}
Hongxiao Bai and Hai Zhao. 2018.
\newblock Deep enhanced representation for implicit discourse relation
  recognition.
\newblock In \emph{COLING}.

\bibitem[{Biran and McKeown(2013)}]{biran2013aggregated}
Or~Biran and Kathleen McKeown. 2013.
\newblock Aggregated word pair features for implicit discourse relation
  disambiguation.
\newblock In \emph{ACL}.

\bibitem[{Bousmalis et~al.(2017)Bousmalis, Silberman, Dohan, Erhan, and
  Krishnan}]{bousmalis17}
Konstantinos Bousmalis, Nathan Silberman, David Dohan, Dumitru Erhan, and Dilip
  Krishnan. 2017.
\newblock Unsupervised pixel-level domain adaptation with generative
  adversarial networks.
\newblock In \emph{CVPR}.

\bibitem[{Braud and Denis(2014)}]{braud14}
Chlo{\'e} Braud and Pascal Denis. 2014.
\newblock Combining natural and artificial examples to improve implicit
  discourse relation identification.
\newblock In \emph{COLING}.

\bibitem[{Braud and Denis(2016)}]{braud16}
Chlo{\'e} Braud and Pascal Denis. 2016.
\newblock Learning connective-based word representations for implicit discourse
  relation identification.
\newblock In \emph{EMNLP}.

\bibitem[{Chen et~al.(2018)Chen, Sun, Athiwaratkun, Cardie, and
  Weinberger}]{Chen2016AdversarialDA}
Xilun Chen, Yu~Sun, Ben Athiwaratkun, Claire Cardie, and Kilian Weinberger.
  2018.
\newblock Adversarial deep averaging networks for cross-lingual sentiment
  classification.
\newblock \emph{Transactions of the Association for Computational Linguistics},
  6:557--570.

\bibitem[{Conneau et~al.(2017)Conneau, Kiela, Schwenk, Barrault, and
  Bordes}]{conneau2017supervised}
Alexis Conneau, Douwe Kiela, Holger Schwenk, Lo{\"i}c Barrault, and Antoine
  Bordes. 2017.
\newblock Supervised learning of universal sentence representations from
  natural language inference data.
\newblock In \emph{EMNLP}.

\bibitem[{Dai and Huang(2018)}]{dai-huang:2018:N18-1}
Zeyu Dai and Ruihong Huang. 2018.
\newblock Improving implicit discourse relation classification by modeling
  inter-dependencies of discourse units in a paragraph.
\newblock In \emph{NAACL}.

\bibitem[{Fu et~al.(2017)Fu, Nguyen, Min, and Grishman}]{Fu2017DomainAF}
Lisheng Fu, Thien~Huu Nguyen, Bonan Min, and Ralph Grishman. 2017.
\newblock Domain adaptation for relation extraction with domain adversarial
  neural network.
\newblock In \emph{IJCNLP}.

\bibitem[{Ganin et~al.(2016)Ganin, Ustinova, Ajakan, Germain, Larochelle,
  Laviolette, Marchand, and Lempitsky}]{Ganinjmlr}
Yaroslav Ganin, Evgeniya Ustinova, Hana Ajakan, Pascal Germain, Hugo
  Larochelle, Fran\c{c}ois Laviolette, Mario Marchand, and Victor Lempitsky.
  2016.
\newblock Domain-adversarial training of neural networks.
\newblock \emph{Journal of Machine Learning Research}, 17(1):2096--2030.

\bibitem[{Goodfellow et~al.(2014)Goodfellow, Pouget-Abadie, Mirza, Xu,
  Warde-Farley, Ozair, Courville, and Bengio}]{goodfellow2014generative}
Ian Goodfellow, Jean Pouget-Abadie, Mehdi Mirza, Bing Xu, David Warde-Farley,
  Sherjil Ozair, Aaron Courville, and Yoshua Bengio. 2014.
\newblock Generative adversarial nets.
\newblock In \emph{NIPS}.

\bibitem[{Gui et~al.(2017)Gui, Zhang, Huang, Peng, and
  Huang}]{Gui2017PartofSpeechTF}
Tao Gui, Qi~Zhang, Haoran Huang, Minlong Peng, and Xuanjing Huang. 2017.
\newblock Part-of-speech tagging for twitter with adversarial neural networks.
\newblock In \emph{EMNLP}.

\bibitem[{Ji et~al.(2015)Ji, Zhang, and Eisenstein}]{ji2015closing}
Yangfeng Ji, Gongbo Zhang, and Jacob Eisenstein. 2015.
\newblock Closing the gap: Domain adaptation from explicit to implicit
  discourse relations.
\newblock In \emph{EMNLP}.

\bibitem[{Joty et~al.(2017)Joty, Nakov, M{\`a}rquez, and
  Jaradat}]{Joty2017CrosslanguageLW}
Shafiq Joty, Preslav Nakov, Llu{\'i}s M{\`a}rquez, and Israa Jaradat. 2017.
\newblock Cross-language learning with adversarial neural networks.
\newblock In \emph{CoNLL}.

\bibitem[{Kingma and Ba(2015)}]{Kingma2014AdamAM}
Diederik~P. Kingma and Jimmy Ba. 2015.
\newblock Adam: A method for stochastic optimization.
\newblock In \emph{ICLR}.

\bibitem[{Lan et~al.(2017)Lan, Wang, Wu, Niu, and Wang}]{Lan2017MultitaskAN}
Man Lan, Jianxiang Wang, Yuanbin Wu, Zheng-Yu Niu, and Haifeng Wang. 2017.
\newblock Multi-task attention-based neural networks for implicit discourse
  relationship representation and identification.
\newblock In \emph{EMNLP}.

\bibitem[{Lan et~al.(2013)Lan, Xu, and Niu}]{lan13}
Man Lan, Yu~Xu, and Zhengyu Niu. 2013.
\newblock Leveraging synthetic discourse data via multi-task learning for
  implicit discourse relation recognition.
\newblock In \emph{ACL}.

\bibitem[{Liu and Tuzel(2016)}]{liu16}
Ming-Yu Liu and Oncel Tuzel. 2016.
\newblock Coupled generative adversarial networks.
\newblock In \emph{NIPS}.

\bibitem[{Liu et~al.(2016)Liu, Li, Zhang, and Sui}]{Liu:2016}
Yang Liu, Sujian Li, Xiaodong Zhang, and Zhifang Sui. 2016.
\newblock Implicit discourse relation classification via multi-task neural
  networks.
\newblock In \emph{AAAI}.

\bibitem[{Marcu and Echihabi(2002)}]{marcu2002unsupervised}
Daniel Marcu and Abdessamad Echihabi. 2002.
\newblock An unsupervised approach to recognizing discourse relations.
\newblock In \emph{ACL}.

\bibitem[{Miltsakaki et~al.(2004)Miltsakaki, Prasad, Joshi, and
  Webber}]{miltsakaki2004penn}
Eleni Miltsakaki, Rashmi Prasad, Aravind Joshi, and Bonnie Webber. 2004.
\newblock {The Penn Discourse Treebank}.
\newblock In \emph{LREC}.

\bibitem[{Miyato et~al.(2018)Miyato, Kataoka, Koyama, and
  Yoshida}]{miyato2018spectral}
Takeru Miyato, Toshiki Kataoka, Masanori Koyama, and Yuichi Yoshida. 2018.
\newblock Spectral normalization for generative adversarial networks.
\newblock In \emph{ICLR}.

\bibitem[{Oza et~al.(2009)Oza, Prasad, Kolachina, Sharma, and
  Joshi}]{oza2009hindi}
Umangi Oza, Rashmi Prasad, Sudheer Kolachina, Dipti~Misra Sharma, and Aravind
  Joshi. 2009.
\newblock {The Hindi Discourse Relation Bank}.
\newblock In \emph{The Third Linguistic Annotation Workshop}.

\bibitem[{Paszke et~al.(2017)Paszke, Gross, Chintala, Chanan, Yang, DeVito,
  Lin, Desmaison, Antiga, and Lerer}]{paszke2017automatic}
Adam Paszke, Sam Gross, Soumith Chintala, Gregory Chanan, Edward Yang, Zachary
  DeVito, Zeming Lin, Alban Desmaison, Luca Antiga, and Adam Lerer. 2017.
\newblock Automatic differentiation in pytorch.
\newblock In \emph{NIPS-W}.

\bibitem[{Pennington et~al.(2014)Pennington, Socher, and
  Manning}]{pennington2014glove}
Jeffrey Pennington, Richard Socher, and Christopher~D. Manning. 2014.
\newblock Glove: Global vectors for word representation.
\newblock In \emph{EMNLP}.

\bibitem[{Pol{\'a}kov{\'a} et~al.(2013)Pol{\'a}kov{\'a}, M{\'\i}rovsk{\`y},
  Nedoluzhko, J{\'\i}nov{\'a}, Zik{\'a}nov{\'a}, and
  Haji{\v{c}}ov{\'a}}]{polakova2013introducing}
Lucie Pol{\'a}kov{\'a}, Ji{\v{r}}{\'\i} M{\'\i}rovsk{\`y}, Anna Nedoluzhko,
  Pavl{\'\i}na J{\'\i}nov{\'a}, {\v{S}}{\'a}rka Zik{\'a}nov{\'a}, and Eva
  Haji{\v{c}}ov{\'a}. 2013.
\newblock Introducing the prague discourse treebank 1.0.
\newblock In \emph{IJCNLP}.

\bibitem[{Prasad et~al.(2008)Prasad, Dinesh, Lee, Miltsakaki, Robaldo, Joshi,
  and Webber}]{Prasad08thepenn}
Rashmi Prasad, Nikhil Dinesh, Alan Lee, Eleni Miltsakaki, Livio Robaldo,
  Aravind Joshi, and Bonnie Webber. 2008.
\newblock {The Penn Discourse TreeBank 2.0}.
\newblock In \emph{LREC}.

\bibitem[{Prasad et~al.(2014)Prasad, Webber, and Joshi}]{prasad2014reflections}
Rashmi Prasad, Bonnie Webber, and Aravind Joshi. 2014.
\newblock Reflections on the {Penn Discourse Treebank}, comparable corpora, and
  complementary annotation.
\newblock \emph{Computational Linguistics}, 40(4):921--950.

\bibitem[{Qin et~al.(2017)Qin, Zhang, Zhao, Hu, and Xing}]{qin17}
Lianhui Qin, Zhisong Zhang, Hai Zhao, Zhiting Hu, and Eric Xing. 2017.
\newblock Adversarial connective-exploiting networks for implicit discourse
  relation classification.
\newblock In \emph{ACL}.

\bibitem[{Russo et~al.(2018)Russo, Carlucci, Tommasi, and Caputo}]{Russo17}
Paolo Russo, Fabio~Maria Carlucci, Tatiana Tommasi, and Barbara Caputo. 2018.
\newblock From source to target and back: symmetric bi-directional adaptive
  gan.
\newblock In \emph{CVPR}.

\bibitem[{Rutherford and Xue(2015)}]{rutherford15}
Attapol Rutherford and Nianwen Xue. 2015.
\newblock Improving the inference of implicit discourse relations via
  classifying explicit discourse connectives.
\newblock In \emph{NAACL}.

\bibitem[{Sankaranarayanan et~al.(2018)Sankaranarayanan, Balaji, Castillo, and
  Chellappa}]{Swami17}
Swami Sankaranarayanan, Yogesh Balaji, Carlos~D. Castillo, and Rama Chellappa.
  2018.
\newblock Generate to adapt: Aligning domains using generative adversarial
  networks.
\newblock In \emph{CVPR}.

\bibitem[{Shi et~al.(2017)Shi, Yung, Rubino, and Demberg}]{shi17}
Wei Shi, Frances Yung, Raphael Rubino, and Vera Demberg. 2017.
\newblock Using explicit discourse connectives in translation for implicit
  discourse relation classification.
\newblock In \emph{IJCNLP}.

\bibitem[{Sporleder and Lascarides(2008)}]{sporleder2008using}
Caroline Sporleder and Alex Lascarides. 2008.
\newblock Using automatically labelled examples to classify rhetorical
  relations: An assessment.
\newblock \emph{Natural Language Engineering}, 14(3):369--416.

\bibitem[{Szegedy et~al.(2016)Szegedy, Vanhoucke, Ioffe, Shlens, and
  Wojna}]{szegedy2016rethinking}
Christian Szegedy, Vincent Vanhoucke, Sergey Ioffe, Jon Shlens, and Zbigniew
  Wojna. 2016.
\newblock Rethinking the inception architecture for computer vision.
\newblock In \emph{CVPR}.

\bibitem[{Tzeng et~al.(2015)Tzeng, Hoffman, Darrell, and Saenko}]{Tzeng15}
Eric Tzeng, Judy Hoffman, Trevor Darrell, and Kate Saenko. 2015.
\newblock Simultaneous deep transfer across domains and tasks.
\newblock In \emph{ICCV}.

\bibitem[{Tzeng et~al.(2017)Tzeng, Hoffman, Saenko, and
  Darrell}]{tzeng2017adversarial}
Eric Tzeng, Judy Hoffman, Kate Saenko, and Trevor Darrell. 2017.
\newblock Adversarial discriminative domain adaptation.
\newblock In \emph{CVPR}.

\bibitem[{Wu et~al.(2016)Wu, Shi, Chen, Huang, and Su}]{wu16}
Changxing Wu, Xiaodong Shi, Yidong Chen, Yanzhou Huang, and Jinsong Su. 2016.
\newblock Bilingually-constrained synthetic data for implicit discourse
  relation recognition.
\newblock In \emph{EMNLP}.

\bibitem[{Wu et~al.(2017)Wu, Shi, Chen, Su, and Wang}]{wu17}
Changxing Wu, Xiaodong Shi, Yidong Chen, Jinsong Su, and Boli Wang. 2017.
\newblock Improving implicit discourse relation recognition with
  discourse-specific word embeddings.
\newblock In \emph{ACL}.

\bibitem[{Xu and Yang(2017)}]{cross17}
Ruochen Xu and Yiming Yang. 2017.
\newblock Cross-lingual distillation for text classification.
\newblock In \emph{ACL}.

\bibitem[{Yang et~al.(2016)Yang, Yang, Dyer, He, Smola, and
  Hovy}]{Yang2016HierarchicalAN}
Zichao Yang, Diyi Yang, Chris Dyer, Xiaodong He, Alexander~J. Smola, and
  Eduard~H. Hovy. 2016.
\newblock Hierarchical attention networks for document classification.
\newblock In \emph{NAACL}.

\bibitem[{Zeyrek et~al.(2013)Zeyrek, Demir{\c{s}}ahin, Sevdik-{\c{C}}all{\i},
  and {\c{C}}ak{\i}c{\i}}]{zeyrek2013turkish}
Deniz Zeyrek, I{\c{s}}{\i}n Demir{\c{s}}ahin, AB~Sevdik-{\c{C}}all{\i}, and
  Ruket {\c{C}}ak{\i}c{\i}. 2013.
\newblock Turkish discourse bank: Porting a discourse annotation style to a
  morphologically rich language.
\newblock \emph{Dialogue and Discourse}, 4(2):174--184.

\bibitem[{Zhang et~al.(2017)Zhang, Barzilay, and Jaakkola}]{aspect}
Yuan Zhang, Regina Barzilay, and Tommi Jaakkola. 2017.
\newblock Aspect-augmented adversarial networks for domain adaptation.
\newblock \emph{Transactions of the Association for Computational Linguistics},
  5:515--528.

\bibitem[{Zhou and Xue(2015)}]{zhou2015chinese}
Yuping Zhou and Nianwen Xue. 2015.
\newblock {The Chinese Discourse TreeBank: a Chinese corpus annotated with
  discourse relations}.
\newblock \emph{Language Resources and Evaluation}, 49(2):397--431.

\bibitem[{Zhou et~al.(2010{\natexlab{a}})Zhou, Lan, Niu, Xu, and Su}]{zhou10_2}
Zhi~Min Zhou, Man Lan, Zheng~Yu Niu, Yu~Xu, and Jian Su. 2010{\natexlab{a}}.
\newblock The effects of discourse connectives prediction on implicit discourse
  relation recognition.
\newblock In \emph{SIGDIAL}.

\bibitem[{Zhou et~al.(2010{\natexlab{b}})Zhou, Xu, Niu, Lan, Su, and
  Tan}]{zhou10}
Zhi-Min Zhou, Yu~Xu, Zheng-Yu Niu, Man Lan, Jian Su, and Chew~Lim Tan.
  2010{\natexlab{b}}.
\newblock Predicting discourse connectives for implicit discourse relation
  recognition.
\newblock In \emph{COLING}.

\end{thebibliography}
\bibliographystyle{acl_natbib}

\end{document}